\documentclass[preprint,12pt]{elsarticle}

\usepackage{enumitem}
\usepackage{tabularx}
\usepackage{subcaption}
\usepackage{notoccite}
\usepackage{afterpage}
\usepackage{threeparttable}
\usepackage{bm}
\usepackage{lmodern}
\usepackage{anyfontsize}
\usepackage{graphicx}%
\usepackage{amsmath,amssymb,amsfonts}%
\usepackage{amsthm}%
\usepackage{mathrsfs}%
\usepackage[title]{appendix}%
\usepackage{xcolor}%
\usepackage{textcomp}%
\usepackage{manyfoot}%
\usepackage{booktabs}%
\usepackage{algorithm}%
\usepackage{algorithmicx}%
\usepackage{algpseudocode}%
\usepackage{listings}%
\usepackage{multirow}%
\let\cline\cmidrule
\usepackage[none]{hyphenat}

\journal{Journal of Pathology Informatics}

\begin{document}
\emergencystretch 3em
\begin{frontmatter}



\title{Contrasting Low and High-Resolution Features for HER2 Scoring using Deep Learning} 

\author[label1]{Ekansh Chauhan\corref{cor1}}
\author[label2]{Anila Sharma\corref{cor1}}
\author[label1]{Amit Sharma}
\author[label2]{Vikas Nishadham}
\author[label2]{Asha Ghughtyal}
\author[label2]{Ankur Kumar}
\author[label2]{Gurudutt Gupta}
\author[label2]{Anurag Mehta}
\author[label1]{C.V. Jawahar}
\author[label3]{P.K. Vinod\corref{cor2}}

\affiliation[label1]{organization={Centre for Visual Information Technology},
            addressline={International Institute of Information Technology},
            city={Hyderabad},
            postcode={500032},
            state={Telangana},
            country={India}}

\affiliation[label2]{organization={Department of Pathology},
            addressline={Rajiv Gandhi Cancer Institute \& Research Centre},
            city={New Delhi},
            postcode={110085},
            state={Delhi},
            country={India}}

\affiliation[label3]{organization={Center for Computational Natural Sciences and Bioinformatics},
            addressline={International Institute of Information Technology},
            city={Hyderabad},
            postcode={500032},
            state={Telangana},
            country={India}}

\cortext[cor1]{Equal Contributions}
\cortext[cor2]{Corresponding author. Email: ekansh.chauhan@research.iiit.ac.in, vinod.pk@iiit.ac.in}

%

\begin{abstract}
Breast cancer, the most common malignancy among women, requires precise detection and classification for effective treatment. Immunohistochemistry (IHC) biomarkers like HER2, ER, and PR are critical for identifying breast cancer subtypes. However, traditional IHC classification relies on pathologists' expertise, making it labor-intensive and subject to significant inter-observer variability. To address these challenges, this study introduces the India Pathology Breast Cancer Dataset (IPD-Breast), comprising of 1,272 IHC slides (HER2, ER, and PR) aimed at automating receptor status classification. The primary focus is on developing predictive models for HER2 3-way classification (0, Low, High) to enhance prognosis. Evaluation of multiple deep learning models revealed that an end-to-end ConvNeXt network utilizing low-resolution IHC images achieved an AUC, F1, and accuracy of 91.79\%, 83.52\%, and 83.56\%, respectively, for 3-way classification, outperforming patch-based methods by over 5.35\% in F1 score. This study highlights the potential of simple yet effective deep learning techniques to significantly improve accuracy and reproducibility in breast cancer classification, supporting their integration into clinical workflows for better patient outcomes.
\end{abstract}


\begin{highlights}
    \item  A breast cancer dataset specific to India (IPD-Breast), was developed to classify HER2, ER, and PR scores from IHC breast cancer slides, addressing a critical gap in computational pathology.
    \item A comprehensive analysis of various approaches for HER2 scoring was performed, including Multiple Instance Learning (MIL), end-to-end CNN-based slide-level classifiers, and patch-level classification pipelines.
    \item The impact of varying magnification levels on the performance of classification methods was evaluated.
    \item Deep learning-based HER2 3-way classification (0, Low, High) using low-resolution whole slide images is proposed, achieving an accuracy of 83.56\%.
\end{highlights}

\begin{keyword}
HER2-low breast cancer\sep immunohistochemistry\sep deep learning\sep digital pathology 


\end{keyword}

\end{frontmatter}



\section{Introduction}\label{sec1}

Breast cancer is a prevalent malignancy that originates from breast cells and can spread if not detected early. It is influenced by genetic, environmental, and lifestyle factors. Early detection through mammography, clinical exams, and advances in histopathological analysis is crucial for better prognosis and survival. Genetic mutations lead to uncontrolled cell growth and tumor formation, with tumors categorized by hormone receptor status, which influences treatment options. Higher incidence rates in developed countries are due to better detection methods, while higher mortality in developing countries is due to limited access to early detection and treatment~\cite{ginsburg2017global}. Immunohistochemistry (IHC) biomarkers like HER2, ER, and PR play a critical role in diagnosis, prognosis, and treatment planning. These biomarkers are essential for personalized therapy, especially for triple-negative breast cancer patients. HER2 is overexpressed in 15-30\% of breast cancers, leading to aggressive growth, with status determined by IHC and in situ hybridization~\cite{godolphin1989studies}. ER and PR receptors, found in 70-80\% of breast cancers, guide hormone therapy treatments~\cite{hammond2010pathologists}. The clinical significance of these biomarkers is summarized in Table \ref{tab_clinical_significance}. 

\begin{table}[ht]
\centering
\begin{tabular}{|c|p{10cm}|}
\hline
\textbf{Biomarker} & \textbf{Clinical Significance} \\
\hline
HER2 0 or 1+ & HER2-negative, no significant  or low overexpression of HER2 protein \\
\hline
HER2 2+ & Equivocal, borderline overexpression; further testing is needed to establish Her2-negative or positive \\
\hline
HER2 3+ & HER2-positive, significant overexpression/amplification of HER2 protein \\
\hline
ER/PR Positive & Less aggressive, better prognosis, responsive to hormone therapies \\
\hline
ER/PR Negative & More aggressive, poorer prognosis, not responsive to hormone therapies \\
\hline
\end{tabular}
\caption{Clinical Significance of HER2, ER, and PR Status}
\label{tab_clinical_significance}
\end{table}

Classification schemes for HER2 status in breast cancer, as illustrated in Table \ref{tab:her2_category}, can be divided into binary, 3-way, and 4-way systems. The binary classification scheme, traditionally used to determine eligibility for HER2-targeted therapies, divides cases into HER2-negative (scores of 0, 1+, and 2+ without ISH amplification) and HER2-positive (scores of 2+ with ISH amplification and 3+)\cite{wolff2018human}. The 3-way classification scheme, gaining prominence due to recent clinical findings from the DESTINY-Breast04 trial, categorizes HER2 status into three groups: 0, Low (IHC 1+ and 2+ without ISH amplification), and High (IHC 3+ or 2+ with ISH amplification)\cite{modi2022trastuzumab, tarantino2020her2}. This scheme recognizes the therapeutic potential for patients with low HER2 expression, who may benefit from new treatments like trastuzumab-deruxtecan~\cite{nakada2019latest}. The 4-way classification scheme, most commonly used, further refines these categories by distinguishing between HER2 0, 1+, 2+, and 3+. The 3-way classification scheme offers significant benefits over other schemes by simplifying the categorization process, making clinical decision-making efficient, and potentially reducing reliance on ISH when the primary goal is to distinguish between low and high HER2 expression for therapeutic decisions.

\begin{table}[ht]
    \centering
    \begin{tabular}{|c|c|c|c|c|c|c|}
        \hline
        \textbf{Task} & \textbf{HER2 Category} & \textbf{0} & \textbf{1+} & \textbf{2+ (ISH -)} & \textbf{2+ (ISH +)} & \textbf{3+} \\
        \hline
        \multirow{2}{*}{Binary} & Negative & \checkmark & \checkmark & \checkmark & & \\
        \cline{2-7}
        & Positive & & & & \checkmark & \checkmark \\
        \hline
        \hline
        \multirow{3}{*}{3-Way} & 0 & \checkmark & &  & & \\
        \cline{2-7}
        & Low & & \checkmark &  \checkmark & & \\
        \cline{2-7}
        & High & & & & \checkmark & \checkmark \\
        \hline
        \hline
        \multirow{4}{*}{4-Way} & 0 & \checkmark & & & & \\
        \cline{2-7}
        & 1+ & & \checkmark & & & \\
        \cline{2-7}
        & 2+ & & & \checkmark & \checkmark & \\
        \cline{2-7}
        & 3+ & & & & & \checkmark \\
        \hline
    \end{tabular}
    \caption{HER2 Classification for Different Tasks. ISH refers to In Situ Hybridization.}
    \label{tab:her2_category}
\end{table}


A significant challenge in breast cancer research is the scarcity of well-annotated datasets that capture the genetic, environmental, and lifestyle diversity of non-Western populations. While most existing datasets are derived from Western cohorts~\cite{datasets1}, which have contributed to the development of computational models and diagnostic tools, there is still a need for region-specific data, particularly from Asia, to ensure the broader applicability and effectiveness of these models in diverse populations. This lack of region-specific data can limit the development of more effective diagnostic and treatment approaches. Compounding this issue is the complexity of IHC subtype classification, which relies on labor-intensive visual slide assessments prone to inter-observer variability and reproducibility issues. The inherent heterogeneity of tumor tissues, with varying levels of marker expression, further complicates accurate classification. Krishnamurthy el al. reported that a mean interobserver variability of 72.4\% among pathologists across all HER2 scores~\cite{krish}. These challenges underscore the critical need for comprehensive, pixel-level annotated datasets, as well as automated, objective, and reproducible computational methods to support pathologists in achieving precise subtype classification.

Machine learning offers significant potential for IHC subtype classification in breast cancer histopathology, specifically for HER2 scoring, which involves classifying patches from whole slide images (WSIs) to predict the HER2 status of the entire slide. Most existing studies rely on patch-level labels, membrane segmentation, or manual region of interest (ROI) selection, which are labor-intensive and challenging to scale. Saha et al. proposed a deep learning framework for semantic segmentation and classification of cell membranes and nuclei to assist in HER2 scoring ~\cite{saha2018her2net}. Cordeiro et al. introduced a method that bypasses segmentation and manual ROI selection, focusing on classifying HER2 scores into four or five categories ~\cite{cordeiro2018automatic}. Kabakıcı et al. proposed a method based on color deconvolution and hybrid multi-level thresholding for segmenting nuclei and boundaries, relying on expert-chosen regions of interest~\cite{kabakcci2021automated}. Wu et el. showed that distinguishing HER2 IHC scores of 0 and 1+ is particularly challenging due to HER2 heterogeneity and observer variability, highlighting a significant disadvantage of manual scoring methods~\cite{WU2023100054}. Tewary et al. introduced AutoIHCNet for HER2 scoring on manually vetted ROIs, with limited patient data~\cite{tewary2022autoihcnet}. A recent study employed DenseNet and ViT models, utilizing small datasets and patch-level annotations~\cite{kabir2024utility}. These methods, while foundational, often involve extensive manual efforts and limited scalability due to data constraints, underscoring the need for more efficient approaches.

ROIs may not capture all relevant information from WSIs and manual annotation of ROIs by pathologists is time consuming. Therefore, WSI-based analysis methods have been proposed to directly predict results for WSIs or patients. Multiple Instance Learning (MIL) is effective for slide-level predictions when pixel-wise annotations are unavailable. Prior studies have demonstrated excellent performance with specific combinations of feature extractors and aggregators using H\&E slides ~\cite{chauhan2024multiple, jiao2024prediction}. MIL's ability to operate at the slide level helps mitigate issues associated with patch-level approaches, which face challenges such as high computational costs and annotation variability. There is a need to develop methods that enable HER2 classification with greater granularity, capturing subtle differences in HER2 expression levels. 


\begin{figure}[h!]
    \centering
    \includegraphics[width=\textwidth]{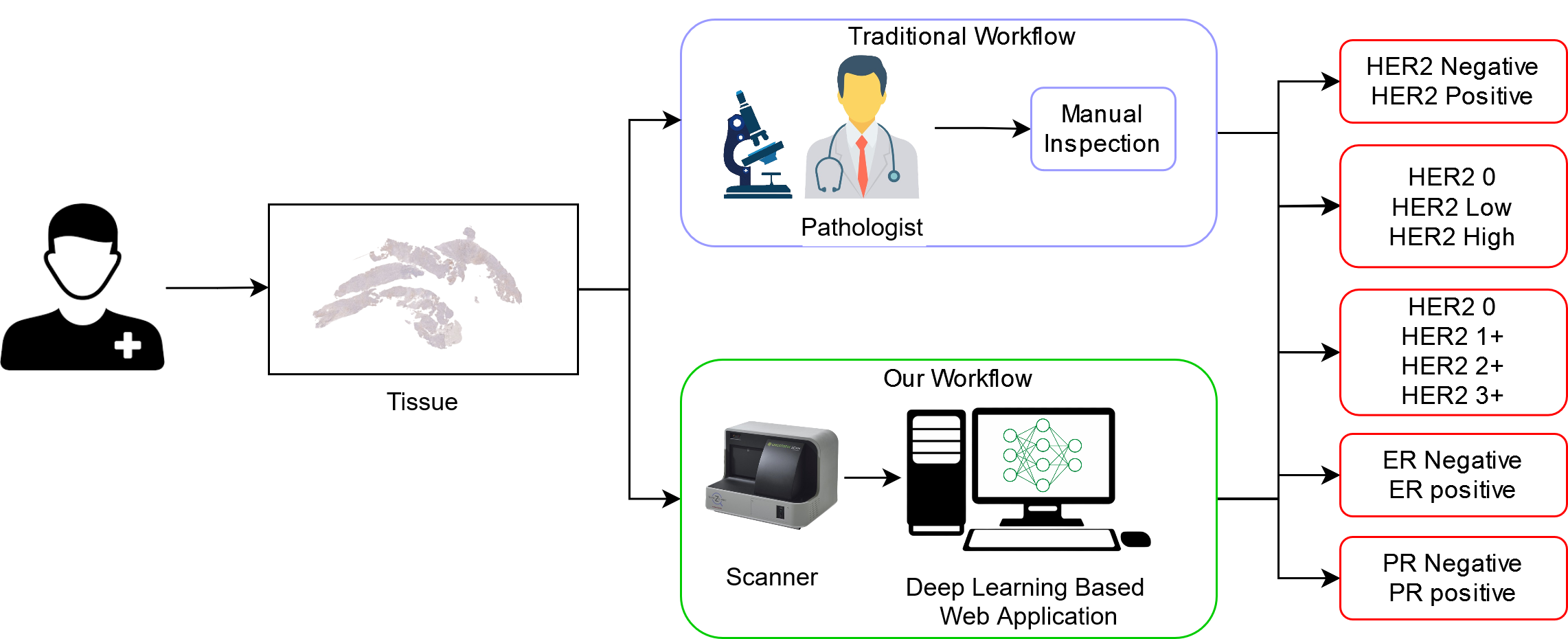}
    \caption{Comparison of traditional manual inspection workflow by pathologists versus a deep learning-based web application for classifying HER2, ER, and PR statuses in breast cancer tissue samples.}
    \label{fig:workflow}
\end{figure}

Given the critical need for accurate breast cancer diagnosis and the limitations of current computational methods, we curated IPD-Breast dataset and evaluated multiple approaches for classifying HER2, ER, and PR statuses from IHC slides. We used slide-level labels to develop end-to-end CNN-based classifiers and MIL-based classifiers, comparing them with patch-level pipelines that use patch-level labels. To the best of our knowledge, this study is the first to apply deep learning on low-resolution WSIs for 3-way HER2 classification. While the MIL-based approach identifies patches effectively, the end-to-end CNN-based method demonstrates superior performance. These models assist pathologists by streamlining evaluations and enhancing diagnostic precision for better patient outcomes (Figure \ref{fig:workflow}).

\section{Materials and Method}

\subsection{Data Acquisition \& Description}

The dataset utilized in this study is part of the larger consortium, IPD India Pathology Dataset, and was retrospectively compiled from the Rajiv Gandhi Cancer Institute and Research Centre (RGCIRC), a premier  tertiary cancer care center in India. It includes a total of 500 patients, with each patient contributing at least two slides (H\&E and HER2), resulting in a total of 2,145 slides. The IPD-Breast dataset includes IHC slides for HER2, ER, PR, and Ki67, as well as H\&E stained slides. Table~\ref{tab:he_her2_count} shows the distribution of 500 IHC slides with varying HER2 expression levels. For ER status, 145 slides are ER-negative and 240 are ER-positive, while for PR status, 198 slides are PR-negative and 189 are PR-positive. The slides were digitized using Hamamatsu's NanoZoomer S210 at 40x magnification, achieving a resolution of 0.23 microns/pixel, ensuring detailed visualization of cellular structures. The study was approved by the RGCIRC institutional review board (Reference Number: RGCIRC/IRB-BHR/75/2023). Additionally, all patient data were anonymized to protect confidentiality, adhering to ethical guidelines for handling clinical data.

Out of 1,000 slides, 764 (both H\&E and HER2) have been annotated (pixel-level) in this phase of data collection. Histopathologists performed detailed annotations using QuPath software, marking specific areas and noting observable features, focusing on primary regions indicative of the disease subtype. These annotations include marking staining presence based on the intensity and the percentage of positively stained cells, following ASCO/CAP guidelines. Each slide underwent a secondary review to ensure accuracy, with discrepancies resolved through discussion or involving a third expert if necessary. 


\begin{table}[ht]
    \centering
    \begin{tabular}{|c|c|c|c|}
        \hline
        \textbf{HER2} & \multicolumn{3}{c|}{\textbf{Count}} \\
        \hline
        0 & \multicolumn{3}{c|}{125} \\
        \hline
        1+ & \multicolumn{3}{c|}{142} \\
        \hline
        \multirow{2}{*}{2+} & \multirow{2}{*}{118} & ISH - & 96 \\
        \cline{3-4}
         & & ISH + & 22 \\
        \hline
        3+ & \multicolumn{3}{c|}{115} \\
        \hline
        \textbf{Total:} & \multicolumn{3}{c|}{\textbf{500}} \\
        \hline
    \end{tabular}
    \caption{Count of HER2 Slides}
    \label{tab:he_her2_count}
\end{table}





\subsection{Methodology}
\subsubsection{MIL Based Classifier}

The analysis for HER2 scoring in histopathology begins with Multiple Instance Learning (MIL) techniques which is well-suited for high-resolution images without patch-level labels. This approach leverages UNI~\cite{chen2024towards} as a feature extractor, a general-purpose self-supervised model pretrained on over 100 million tissue patches using Vision Transformer architecture and DINOv2 self-supervised learning. UNI captures high-level visual features but faces limitations in IHC applications due to its H\&E training background and the specific requirements for HER2 scoring, which involves different staining techniques and precise interpretation of cell membrane patterns. We evaluate two feature aggregation methods, CLAM~\cite{lu2021data} and DTFD~\cite{Zhang_2022_CVPR}, for HER2 scoring classification.

CLAM evaluates and ranks patches within a whole slide by assigning attention scores based on their significance. It aggregates these scores to form a slide-level representation for classification, with a single-branch configuration. DTFD uses pseudo-bags by dividing instances into smaller bags to refine attention-based MIL models. This approach improves feature extraction and prediction accuracy. In this study, we use the Aggregated Feature Selection (AFS) strategy from DTFD to enhance classification performance despite potential label noise.

The MIL framework for HER2 classification faces several challenges in meeting ASCO/CAP guidelines. Mixed subtypes within a single slide can obscure minor subtypes, complicating accurate classification. Additionally, the guideline requiring at least 10\% of tumor cells of the labeled category on each slide presents a challenge, as traditional MIL struggles with aggregating dispersed information across patches. Label ambiguity in heterogeneous tumor slides further complicates the process, as MIL assumes clear labels for each bag. While attention mechanisms can help prioritize critical instances, they may not fully capture the spatial and proportional nuances required for accurate HER2 scoring. Therefore, a modified approach incorporating spatial relationships is needed for reliable HER2 classification.

\subsubsection{End-to-End Slide Level Classifier}

To address these challenges, we adopted Convolutional Neural Networks (CNNs), which are well-suited for analyzing spatial data and detecting intricate patterns within images. For this task, we utilized the ConvNeXt architecture~\cite{liu2022convnet}, a modern CNN model known for its high performance in image classification tasks. ConvNeXt incorporates several advancements over traditional CNNs, including depthwise separable convolutions that reduce computational complexity while maintaining performance, and residual connections that enhance training stability and convergence. It also features inverted bottlenecks that expand the hidden dimension in the Multi-Layer Perceptron (MLP) block, inspired by Transformer models, to improve feature extraction and overall model performance.

ConvNeXt employs Group Normalization instead of Batch Normalization, which is particularly effective with small batch sizes and improves training stability. The model uses Gaussian Error Linear Units (GELU) for smoother activation functions compared to Rectified Linear Units (ReLU). It also integrates larger kernel sizes, such as 7x7, to capture more global context, and incorporates design elements from Vision Transformers, including hierarchical structures and a patch-based approach to process high-resolution images efficiently. Training methodologies include the AdamW optimizer, learning rate scheduling, data augmentation, and extensive hyperparameter tuning, which together contribute to ConvNeXt's state-of-the-art performance on image classification benchmarks. This approach provides a robust end-to-end framework for accurate HER2 classification by effectively capturing and analyzing complex histopathological features.

\subsubsection{Patch-Level Classification Pipeline}

End-to-end Convolutional Neural Networks (CNNs) have proven effective for generalization with limited data but face significant challenges when applied to HER2 classification using low-resolution images. One major issue is the loss of crucial morphological and color details, which are essential for accurate HER2 assessment. Low-resolution images often fail to capture the fine details necessary for precise classification, potentially degrading model performance. Additionally, the reliance on slide-level labels introduces inconsistencies due to inter-observer variability, which can affect model accuracy and reliability, leading to biased or inconsistent data.

To overcome these limitations, we shift our focus from slide-level to patch-level classification using the ConvNeXt model, which has previously demonstrated strong performance at the slide level. This approach involves creating a dataset of image patches from slides with partial annotations. Each patch is labeled based on the inherited annotations from the original slide, allowing for the development of a patch-level classifier. During inference, patches are classified to determine HER2 expression across the entire slide. Aggregated patch-level HER2 scores are then used by machine learning classifiers (Random Forest, SVM, and MLP) to make the final slide-level HER2 classification.

The patch-level classification approach enhances analysis granularity by focusing on smaller slide regions, improving diagnostic precision for HER2 expression. By training the ConvNeXt model with patch-level labels, the method provides a detailed examination of HER2 expression, capturing variability across different patches. This detailed patch-level analysis aligns with clinical realities of intra-tumor heterogeneity, resulting in more accurate and clinically relevant HER2 scoring. This method ensures that the model accounts for the variability in marker expression within a tumor, enhancing both its diagnostic accuracy and applicability.


\subsection{Experiment Setup}

For pre-processing, we started with foreground and background segmentation of whole slide images (WSIs) using contrast enhancement and Gaussian blurring. Otsu's thresholding and watershed segmentation were applied to isolate the background. Patches of size 224x224 were then extracted at high resolution using the CLAM pipeline and processed with a pre-trained UNI model. Different feature aggregators are evaluated, with the data split into training, validation, and test sets on a patient-wise basis (80/10/10 split). Classification tasks include binary, three-way, and four-way categorizations.

For the CNN-based classifiers, including ResNet-50, DenseNet-201, ViT-Base, and ConvNeXt-Small, low-resolution images (224x224 and 512x512) are used, with 10-fold stratified cross-validation to assess model robustness. Models are trained for 25 to 100 epochs with early stopping based on validation loss, using the Adam optimizer and various augmentations such as flips, rotations, and affine transformations. Cross-entropy, class weights, and focal loss functions are employed for multi-class tasks, and the best-performing model is selected for ER and PR binary classification.

Furthermore, for patch-level classification, a patch-level dataset is created from slide-level data with partial annotations. Patches are extracted at multiple resolutions and checked for significant overlaps with annotated regions using a minimum intersection threshold of 5\%. White and black patches are excluded based on pixel ratio thresholds (70\% and 20\% respectively), and Contrast Limited Adaptive Histogram Equalization (CLAHE) is applied for quality control. After these steps, the patches that passed all checks were saved to specific directories based on their resolution level. Each patch was labeled according to the intersecting annotation, ensuring accurate and informative labeling.

\section{Results}

The performance of the various methods for HER2 classification using different machine learning approaches was evaluated across multiple classification tasks: binary, 3-way, and 4-way. The results provide insights into the strengths and weaknesses of each approach and highlight their respective capabilities in handling the complexities of histopathological image analysis. The MIL-based classifiers, specifically CLAM-sb, CLAM-mb, and DTFD, demonstrated competitive performance across all classification tasks (supplementary table~\ref{tab:MIL_results}). DTFD outperforms others in terms of accuracy and F1 score for 4-way classification, while CLAM-sb excels in the 3-way classification task.


\begin{table}[!h]
    \centering
    \begin{tabular}{|c|c|c|c|c|}
        \hline
        \textbf{Task} & \textbf{Method} & \textbf{AUC (\%)} & \textbf{F1 (\%)} & \textbf{Accuracy (\%)} \\
        \hline
        \multirow{3}{*}{Binary} & ResNet-50 & 95.70 $\pm$ 2.79 & 85.44 $\pm$ 7.56 & 92.80 $\pm$ 3.12  \\
        \cline{2-5}
         & DenseNet-201 & 95.73 $\pm$ 1.62 & 87.84 $\pm$ 4.79 & 93.60 $\pm$ 2.33  \\
        \cline{2-5}
         & \textbf{ConvNext-S} & \textbf{96.15 $\pm$ 3.22} & \textbf{92.75 $\pm$ 5.37} & \textbf{96.19 $\pm$ 2.54} \\
        \cline{2-5}
         & ViT-B & 95.57 $\pm$ 3.55 & 87.39 $\pm$ 8.95 & 93.80 $\pm$ 4.14  \\
        \hline
        \hline
        \multirow{3}{*}{3-Way} & ResNet-50  & 86.95 $\pm$ 2.72 & 70.81 $\pm$ 6.44 & 72.20 $\pm$ 5.01  \\
        \cline{2-5}
         & DenseNet-201 &  87.64 $\pm$ 3.79 & 73.22 $\pm$ 6.06 & 74.2 $\pm$ 6.09 \\
        \cline{2-5}
         & \textbf{ConvNext-S}  & \textbf{91.79 $\pm$ 2.14} & \textbf{83.52 $\pm$ 2.69} & \textbf{83.56 $\pm$ 2.80} \\
        \cline{2-5}
         & ViT-B & 84.45 $\pm$ 4.66 & 68.32 $\pm$ 6.01 & 69.00 $\pm$ 6.21 \\
        \hline
        \hline
        \multirow{3}{*}{4-Way} & ResNet-50 & 86.45 $\pm$ 3.00 & 63.08 $\pm$ 4.80 & 62.60 $\pm$ 4.29  \\
        \cline{2-5}
         & DenseNet-201 & 84.83 $\pm$ 3.71 & 60.73 $\pm$ 6.43 & 61.20 $\pm$ 5.74 \\
        \cline{2-5}
         & \textbf{ConvNext-S} & \textbf{89.97 $\pm$ 2.56} & \textbf{75.61 $\pm$ 4.42} & \textbf{75.53 $\pm$ 4.02} \\
        \cline{2-5}
         & ViT-B & 83.64 $\pm$ 3.99 & 60.47 $\pm$ 9.75 & 60.20 $\pm$ 9.14 \\
        \hline
    \end{tabular}
    \caption{Approach-2 (End-to-End CNN) Results: 10-Fold Performance of Various CNN-Based Classifiers on Test Set for Slide-Level Classification Tasks.}
    \label{tab:cnn_metric}
\end{table}

\begin{figure}[h!]
    \centering
    \includegraphics[width=0.6\textwidth]{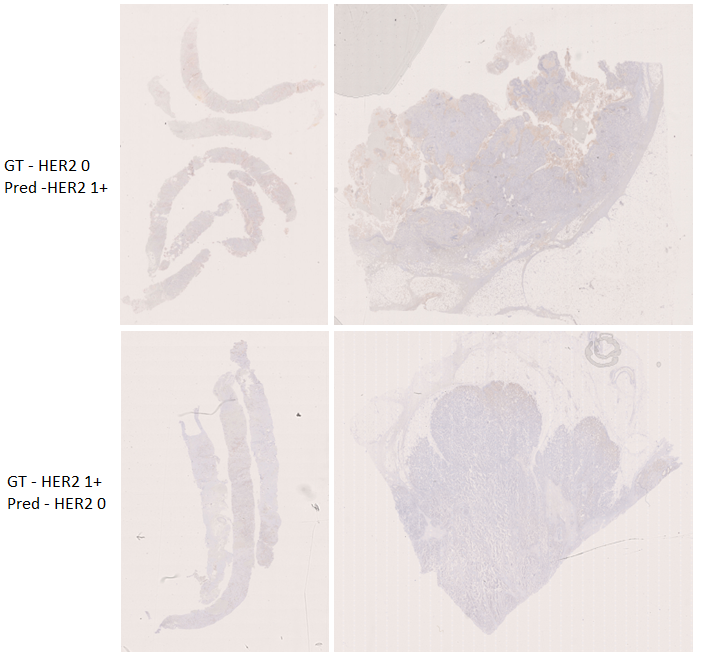}
    \caption{Few misclassified samples in Approach 2. [GT- Ground Truth, Pred- Predicted class]}
    \label{fig:compare_a2}
\end{figure}

Table~\ref{tab:cnn_metric} depicts the performance of Approach-2 (CNN-based classifiers) including ResNet-50, DenseNet-201, ViT-B, and ConvNext-S, each showing different levels of performance across the classification tasks. ConvNext-S emerges as the top performer among them, achieving best metrics across binary 3-way, and 4-way classification tasks. Its higher AUC, F1, and accuracy scores compared to ResNet-50 and DenseNet-201 highlight its enhanced capability in feature extraction and classification. The performance of ConvNeXt-S can be attributed to its ability to maintain a high level of context, which is essential for accurate HER2 scoring. Figure~\ref{fig:compare_a2} shows the misclassified samples of HER2 0 and HER2 1+. It can be observed that the confusion between these classes arises due to discrepancies between the image features (intensity and percentage of brown color) and corresponding labels. This can be attributed to the high inter-observer variability inherent in manual HER2 scoring.

\begin{table}[ht]
    \centering
    \begin{tabular}{|c|c|c|c|c|}
        \hline
        \textbf{Average Resolution} & \textbf{Total Patches} &\textbf{AUC (\%)} & \textbf{F1 (\%)} & \textbf{Accuracy (\%)} \\
        \hline
        Low (1648 * 1471) & 2155 & 94.29 $\pm$ 2.25 & 77.85 $\pm$ 6.07 & 78.83 $\pm$ 5.81 \\
        \hline
        Moderate \textbf{(4549 * 4017)} & \textbf{11065} & \textbf{94.75 $\pm$ 1.65 }& \textbf{78.25 $\pm$ 4.67} & \textbf{81.43 $\pm$ 4.24} \\
        \hline
        High (23218 * 20529) & 198807 & 94.37 $\pm$ 1.89 & 77.56 $\pm$ 4.08 & 80.55 $\pm$ 3.92 \\
        \hline
    \end{tabular}
    \caption{10-Fold Performance of ConvNext on Test Set (Patch Dataset) for 4-way Patch-Level Classification on various resolutions.}
    \label{tab:patch_vary_res}
\end{table}

\begin{figure}[h!]
    \centering
    \includegraphics[width=0.8\textwidth]{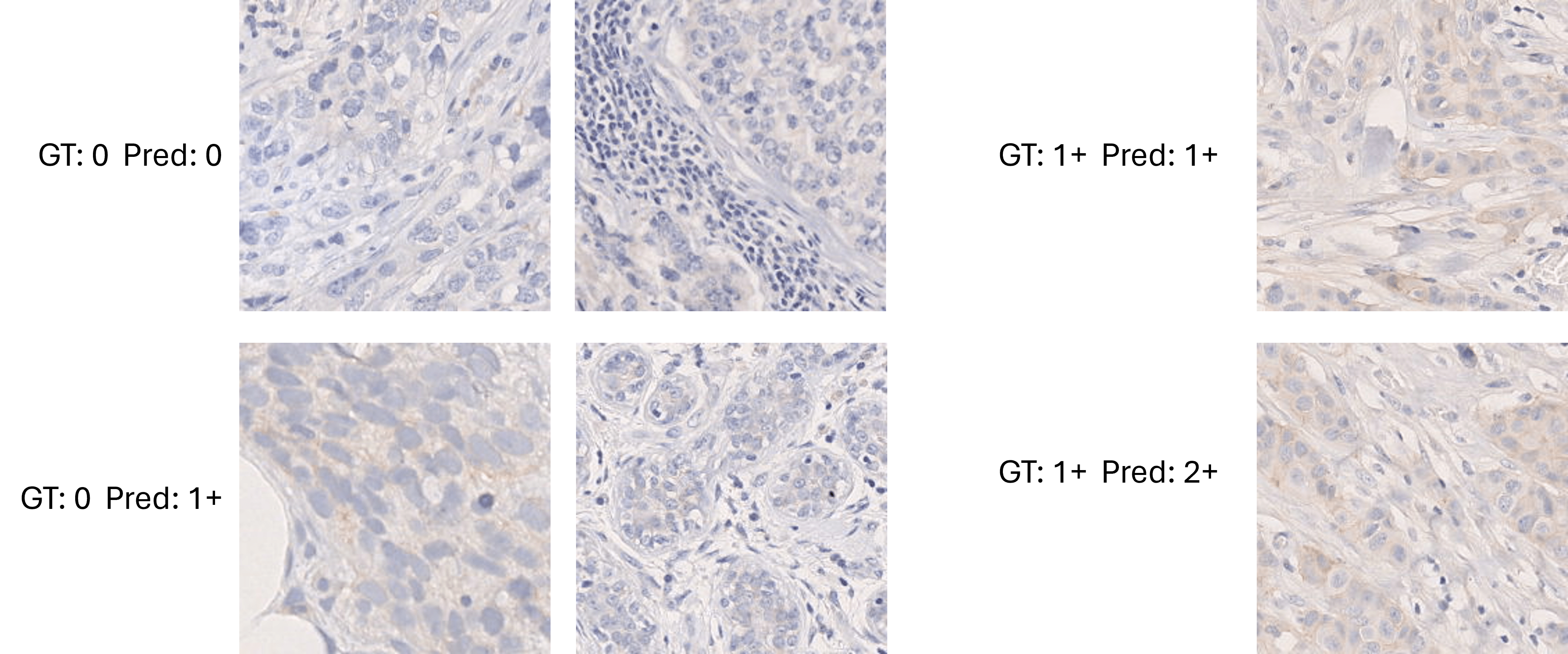}
    \caption{Few predicted samples at patch level using Approach 3. Top row shows correctly predicted samples whereas bottom row shows the misclassified samples. [GT- Ground Truth, Pred- Predicted class]}
    \label{fig:compare_a3}
\end{figure}

The patch-level classifier approach, evaluated at low (average resolution: 1,648x1,471), moderate (average resolution: 4,549x4,017), and high (average resolution: 23,218x20,529) resolutions, highlights the importance of granularity in HER2 classification. Table~\ref{tab:patch_vary_res} shows that moderate resolution setting achieved the best performance, balancing detail and computational efficiency with an AUC of 94.75, an F1 score of 78.25, and an accuracy of 81.43. High-resolution patches, while detailed, did not offer performance gains and incurred higher computational costs. Low-resolution patches underperformed due to insufficient detail. This approach enhances diagnostic precision by focusing on smaller regions within slides, effectively handling tumor heterogeneity and reducing overfitting through a larger number of training samples. Despite higher computational demands and the need for precise annotations, the patch-level method shows robust performance for improving HER2 classification accuracy in clinical practice. Figure~\ref{fig:compare_a3} shows the misclassified patch level samples. It can be observed that variability exists in the annotated tissue regions, as a single annotated region may contain patches belonging to multiple HER2 classes.

\begin{table}[ht]
    \centering
    \begin{tabular}{|c|c|c|c|c|}
        \hline
        \textbf{Task} & \textbf{Method} & \textbf{AUC (\%)} & \textbf{F1 (\%)} & \textbf{Accuracy (\%)} \\
        \hline
        \multirow{3}{*}{Binary} & Random Forest & 97.37 $\pm$ 2.80 & 91.35 $\pm$ 4.89 & 93.15 $\pm$ 3.82 \\
        \cline{2-5}
         & \textbf{SVM} & 96.70 $\pm$ 3.72 & \textbf{91.81 $\pm$ 4.47} & 93.50 $\pm$ 3.62 \\
        \cline{2-5}
         & MLP & \textbf{97.82 $\pm$ 1.88} & 91.64 $\pm$ 3.81 & \textbf{93.52 $\pm$ 3.26} \\
        \hline
        \hline
        \multirow{3}{*}{3-Way} & Random Forest & 91.12 $\pm$ 2.42 & 76.03 $\pm$ 2.50 & 77.63 $\pm$ 3.64 \\
        \cline{2-5}
         & \textbf{SVM} & 90.84 $\pm$ 2.96 & \textbf{78.17 $\pm$ 4.30} & \textbf{79.11 $\pm$ 5.22} \\
        \cline{2-5}
         & MLP & \textbf{92.55 $\pm$ 2.21} & 77.14 $\pm$ 3.87 & 78.18 $\pm$ 5.11 \\
        \hline
        \hline
        \multirow{3}{*}{4-Way} & Majority Voting & 42.77 & 13.88 & 16.66 \\
        \cline{2-5}
         & Random Forest & 91.07 $\pm$ 2.90  & 71.73 $\pm$ 3.73 & 72.38 $\pm$ 4.37 \\
        \cline{2-5}
         & SVM & 91.53 $\pm$ 2.14 & 73.07 $\pm$ 5.26 & 73.46 $\pm$ 5.72 \\
        \cline{2-5}
         & \textbf{MLP} & \textbf{91.78 $\pm$ 2.54} & \textbf{73.49 $\pm$ 5.07} & \textbf{73.81 $\pm$ 5.60} \\
        \hline
    \end{tabular}
    \caption{Approach-3 (Patch-Based) results at \textbf{low Resolution}: 10-fold performance of various machine learning classifiers for slide-level classification tasks using count of HER2 score patches.}
    \label{tab:low_res_classifier}
\end{table}

The distribution-based slide-level classifier (Approach-3) was evaluated using Random Forest, SVM, and MLP classifiers along with majority voting across three resolutions: Low, Moderate, and High. The performance metrics, including AUC, F1 score, and accuracy, revealed that low resolution exhibits strong performance across all tasks (Table \ref{tab:low_res_classifier}). Notably, MLP achieved highest AUC for all the tasks whereas SVM achieved higher F1 score for binary and 3-way. Moderate resolution also performed comparably to low resolution (Supplementary Table \ref{tab:high_res_classifier}), with Random Forest and MLP showing better performance. However, high resolution did not provide improvements over low and moderate resolutions (Supplementary Table \ref{tab:very_high_res_classifier}). For binary classification, MLP achieved the highest performance with an AUC of 96.67, F1 score of 88.52, and accuracy of 91.78.

Although the models demonstrated highly competitive performance across resolutions, low-resolution images surprisingly yielded better or comparable metrics to moderate and high resolutions. This indicates that essential features can be effectively captured even at lower resolutions, enabling accurate diagnosis while reducing computational costs. Moderate resolution offers a balanced approach, providing a trade-off between detail and efficiency, whereas high resolution often fails to provide significant benefits and may introduce noise and complexity without substantial gains. These results underscore the importance of carefully selecting the resolution and classifier to balance performance and computational efficiency in HER2 classification tasks.


\begin{figure}[ht]
    \centering
    \begin{subfigure}[b]{0.48\textwidth}
        \centering
        \includegraphics[width=\textwidth]{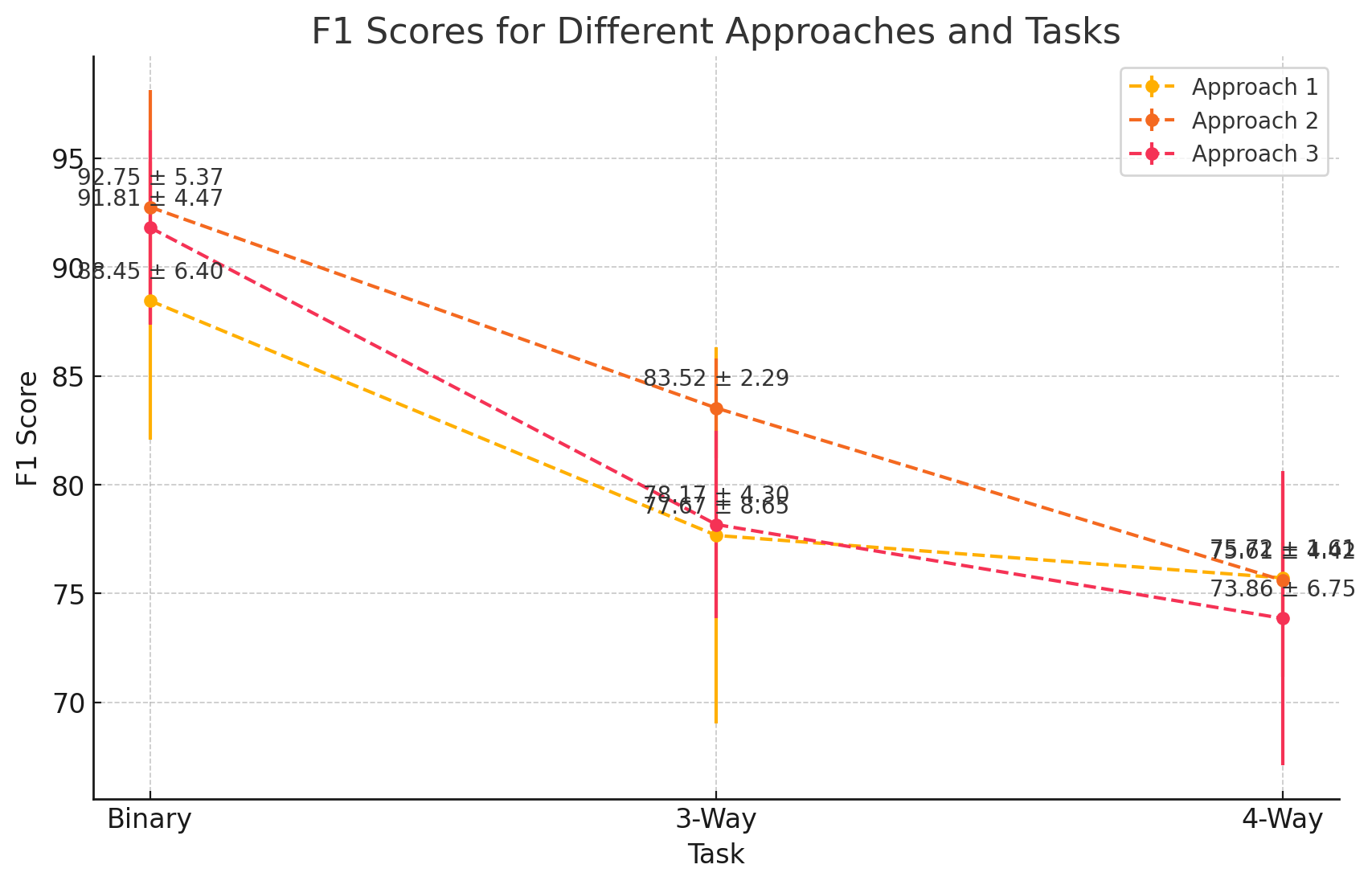}
        \caption{}
        \label{fig:4subfig1}
    \end{subfigure}
    \hfill
    \begin{subfigure}[b]{0.48\textwidth}
        \centering
        \includegraphics[width=\textwidth]{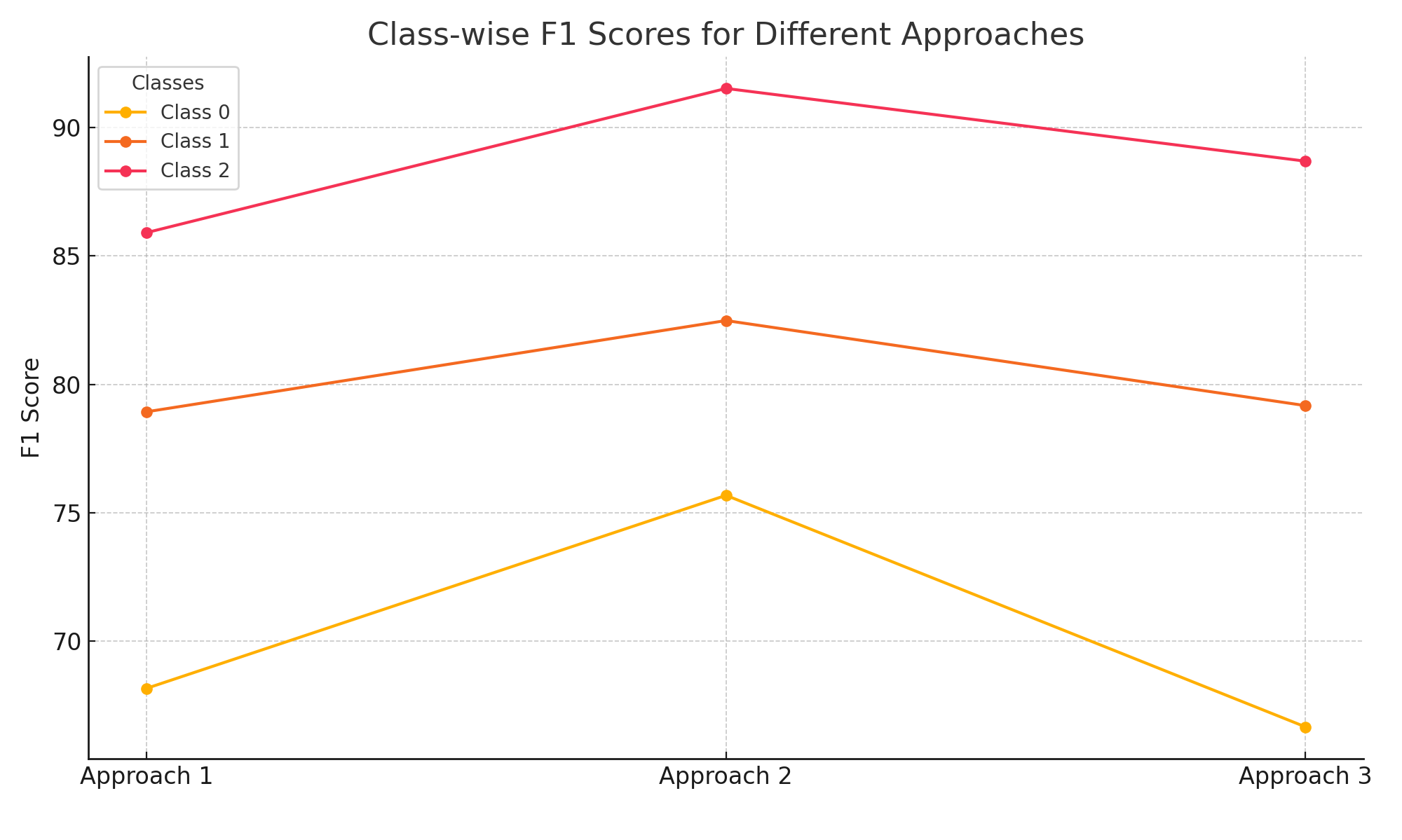}
        \caption{}
        \label{fig:4subfig2}
    \end{subfigure}
    \caption{Comparison of F1 scores for different approaches. (a) Overall performance across tasks. (b) Class-wise performance in the 3-Way classification task.}
    \label{fig:4combined}
\end{figure}

A comparison of the best models from each approach (Figure \ref{fig:4subfig1}), reveals that Approach 2, featuring the end-to-end ConvNeXt-S model, delivers the most robust performance across all tasks.  For the 3-way classification task, which is the primary focus of this study, Approach 2 achieves the highest AUC of 91.79 and an F1 score of 83.52, underscoring its effectiveness. In binary classification, Approach 2 outperforms the others with an impressive F1 score of 92.75, demonstrating its ability to distinguish between HER2-positive and HER2-negative cases. For the 4-way classification task, although Approach 1 (MIL-based) slightly outperforms the others with an AUC of 92.79, the performance differences were minimal, indicating that ConvNeXt-S maintains strong competitive performance even in more complex classification tasks.


Class-wise analysis also provides valuable insights into the strengths and weaknesses of each method (Figure \ref{fig:4subfig2}). This show that Approach 2 (ConvNeXt-S) achieves a higher F1 score for each class compared to other approaches. A notable performance drop is observed in Class 0 (HER2 0) and Class 1 (HER2 1+) across all approaches. This decline can be attributed to subtle differences in staining patterns between HER2 0 and HER2 1+, which also contribute to inter- and intra-observer variability in slide labeling. These findings highlight the difficulty in accurately classifying intermediate classes and underscore the need for models that can more effectively differentiate subtle variations in histopathological features.

\begin{figure}[ht]
    \centering
    \begin{subfigure}{0.45\textwidth}
        \centering
        \includegraphics[width=\linewidth]{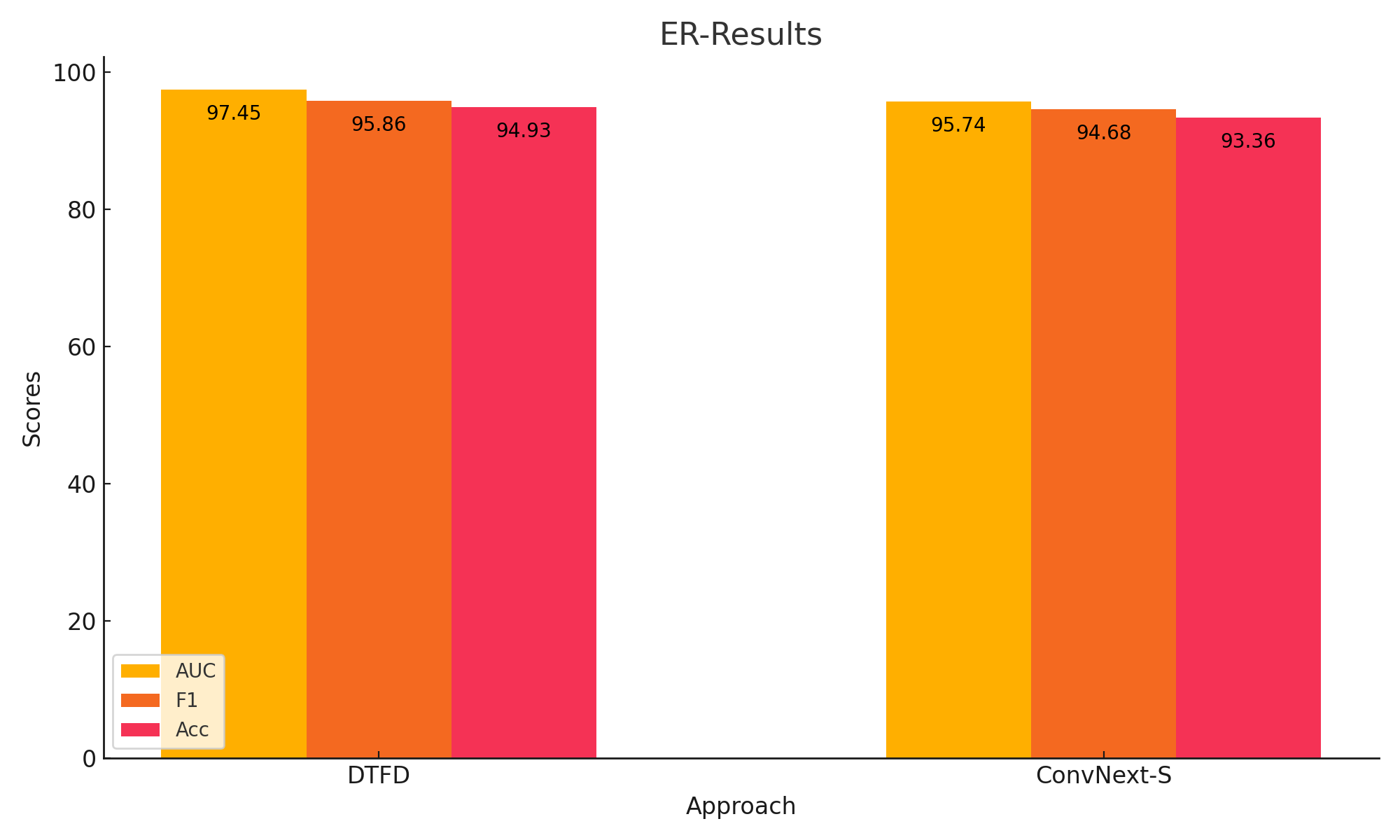}
        \caption{}
    \end{subfigure}
    \hfill
    \begin{subfigure}{0.45\textwidth}
        \centering
        \includegraphics[width=\linewidth]{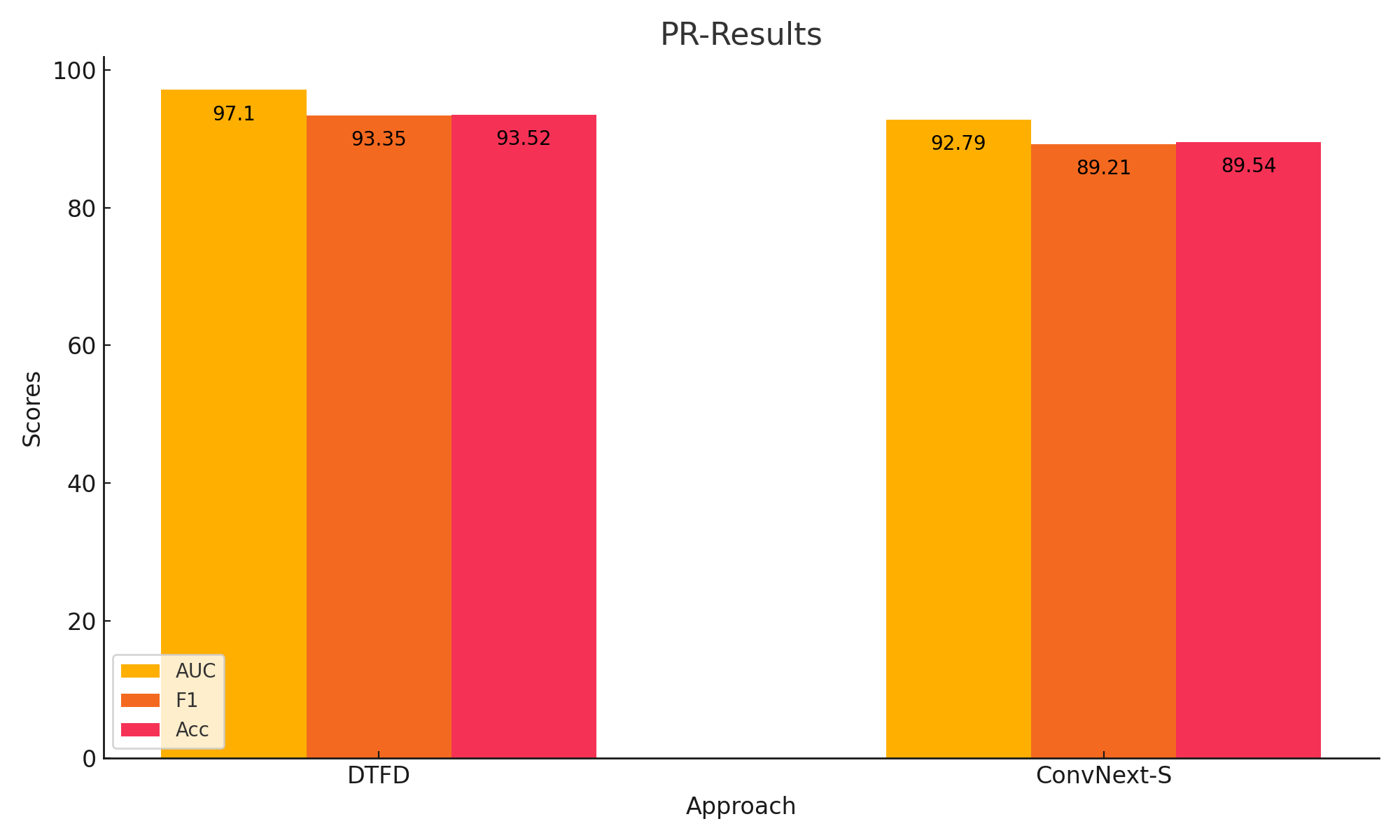}
        \caption{}
    \end{subfigure}
    \caption{Comparison of DTFD and ConvNext-S approaches for (a) ER and (b) PR binary classification}
    \label{fig:er-pr-results}
\end{figure}

We also attempted to classify Estrogen Receptor (ER) and Progesterone Receptor (PR) statuses using IHC slides. Figure \ref{fig:er-pr-results} shows the comparison of DTFD and ConvNeXt-S results for PR and ER classification tasks. For PR classification, DTFD outperforms ConvNeXt-S with an AUC of 97.1 and an F1 score of 93.35. Similarly, in ER classification, DTFD leads with an AUC of 97.45 and an F1 score of 95.86. Although ConvNeXt-S slightly trails, it still demonstrates strong performance with an AUC of 92.79 and an F1 score of 89.21 for PR, and an AUC of 95.74 and an F1 score of 94.68 for ER. These results suggest that, while both methods are effective, DTFD provides higher accuracy and reliability in ER and PR status classification tasks.


\section{Discussion}

This study explored various approaches for HER2 scoring using artificial intelligence, each offering unique strengths and challenges. A key observation from the experiments is that increasing resolution of IHC slides for training did not result in significant improvements in either patch-level or slide-level performance. This indicates that while resolution of patches is important, it may not be the primary factor limiting performance for this task. Instead, aggregation of patch-level predictions during inference appears to be more impactful, as effective aggregation can better capture the spatial context and heterogeneity within WSIs, providing a more accurate representation of HER2 expression at the slide level.

Among the approaches evaluated, Approach-2, based on the ConvNeXt architecture, emerged as the most effective. ConvNeXt’s advanced convolutional structure, including depthwise separable convolutions and residual connections, allowed it to effectively model intricate spatial patterns and staining variations in histopathological images. Its ability to process WSIs holistically helped to preserve the spatial context, which is essential for tasks like HER2 scoring, where cell distribution and stain intensity are key factors.

Although Approach-2 demonstrated superior performance, Approaches 1 and 3 provided enhanced clinical interpretability. Approach-1 uses attention mechanisms to emphasize critical regions within WSIs, ensuring that the model focuses on clinically significant areas. Approach-3 provided a more detailed picture by classifying individual patches, aligning with clinical workflows that often focus on specific regions of interest. In contrast, while Approach-2's low-resolution design proved effective for holistic classification, it lacked the fine-grained interpretability.

The binary and 3-way classification datasets were derived from an initially balanced 4-way classification dataset, which introduced class imbalance. To mitigate this issue, various techniques were employed, including focal loss, class weighting, and data augmentation. Focal loss was applied to place greater emphasis on misclassified and minority-class samples, while class weighting adjusted the loss function according to class frequencies. Data augmentation was used to artificially enhance the representation of minority classes by diversifying the dataset. However, despite these efforts, none of these methods led to significant improvements in classification outcomes.

 Most HER2 scoring approaches relied on annotated training data and multiple computational steps, including identifying tumor cells, classifying staining patterns based on membrane intensity and completeness, and deriving slide-level scores from cell counts. In contrast, we present a simplified solution using slide-level annotations and one of the largest available HER2 scoring datasets. Despite this, ConvNeXt-S achieves superior performance, with an accuracy exceeding 83\% in the 3-way classification task.
 
 A recent study demonstrated that the AI solution achieved high accuracy for HER2 scoring, with 92.1\% agreement on slides with high-confidence ground truth, using only a smaller dataset~\cite{krish}. There is a need to evaluate the concordance of our model with high-confidence ground truth, as we cannot exclude the potential impact of inter-observer variability in our ground truth (Figure~\ref{fig:compare_a2}). While the models demonstrate promising capabilities, further validation and refinement are necessary to ensure their reliability and integration into clinical workflows. To facilitate this process, an API based on this model will be deployed at the hospital for validation and iterative improvement. Further variations in staining procedures and images generated by different scanners can impact performance. Efforts will focus on addressing current limitations and improving models to ensure practical utility in real-world scenarios. These advancements hold the potential to enhance the accuracy and consistency of HER2 classification, thereby supporting pathologists in delivering better patient care.

\section*{Declaration of competing interest}
All authors declare they have no conflicts of interest.
\section*{Acknowledgments}
This work was supported by iHUB-Data, International Institute of Information Technology, Hyderabad, India.
\section*{Data Availability}
The data that has been used is confidential.

\bibliography{cas-refs}

\begin{thebibliography}{10}

\bibitem{ginsburg2017global}
Ophira Ginsburg, Freddie Bray, Michel~P Coleman, Verna Vanderpuye, Alexandru
  Eniu, S~Rani Kotha, Malabika Sarker, Tran~Thanh Huong, Claudia Allemani,
  Allison Dvaladze, et~al.
\newblock The global burden of women’s cancers: a grand challenge in global
  health.
\newblock {\em The Lancet}, 389(10071):847--860, 2017.

\bibitem{godolphin1989studies}
William Godolphin, Lovell~A Jones, John~A Holt, Steven~G Wong, Duane~E Keith,
  W~Levin, S~Stuart, Judy Udove, Axel Ullrich, et~al.
\newblock Studies of the her-2/neu protooncogene in human breast and ovarian
  cancer.
\newblock {\em Science}, 244(4905):707--712, 1989.

\bibitem{hammond2010pathologists}
ME~Hammond, DF~Hayes, M~Dowsett, DC~Allred, KL~Hagerty, S~Badve, et~al.
\newblock Pathologists’ guideline recommendations for immunohistochemical
  testing of estrogen and progesterone receptors in breast cancer.
\newblock {\em Breast Care}, 5(3):185--7, 2010.

\bibitem{wolff2018human}
Antonio~C Wolff, M~Elizabeth~Hale Hammond, Kimberly~H Allison, Brittany~E
  Harvey, Pamela~B Mangu, John~MS Bartlett, Michael Bilous, Ian~O Ellis,
  Patrick Fitzgibbons, Wedad Hanna, et~al.
\newblock Human epidermal growth factor receptor 2 testing in breast cancer:
  American society of clinical oncology/college of american pathologists
  clinical practice guideline focused update.
\newblock {\em Archives of pathology \& laboratory medicine},
  142(11):1364--1382, 2018.

\bibitem{modi2022trastuzumab}
Shanu Modi, William Jacot, Toshinari Yamashita, Joohyuk Sohn, Maria Vidal,
  Eriko Tokunaga, Junji Tsurutani, Naoto~T Ueno, Aleix Prat, Yee~Soo Chae,
  et~al.
\newblock Trastuzumab deruxtecan in previously treated her2-low advanced breast
  cancer.
\newblock {\em New England Journal of Medicine}, 387(1):9--20, 2022.

\bibitem{tarantino2020her2}
Paolo Tarantino, Erika Hamilton, Sara~M Tolaney, Javier Cortes, Stefania
  Morganti, Emanuela Ferraro, Antonio Marra, Giulia Viale, Dario Trapani,
  Fatima Cardoso, et~al.
\newblock Her2-low breast cancer: pathological and clinical landscape.
\newblock {\em Journal of Clinical Oncology}, 38(17):1951--1962, 2020.

\bibitem{nakada2019latest}
Takashi Nakada, Kiyoshi Sugihara, Takahiro Jikoh, Yuki Abe, and Toshinori
  Agatsuma.
\newblock The latest research and development into the antibody--drug
  conjugate,[fam-] trastuzumab deruxtecan (ds-8201a), for her2 cancer therapy.
\newblock {\em Chemical and Pharmaceutical Bulletin}, 67(3):173--185, 2019.

\bibitem{datasets1}
Talha Qaiser, Abhik Mukherjee, Chaitanya Reddy~Pb, Sai~D Munugoti, Vamsi
  Tallam, Tomi Pitk{\"a}aho, Taina Lehtim{\"a}ki, Thomas Naughton, Matt
  Berseth, An{\'\i}bal Pedraza, et~al.
\newblock Her 2 challenge contest: a detailed assessment of automated her 2
  scoring algorithms in whole slide images of breast cancer tissues.
\newblock {\em Histopathology}, 72(2):227--238, 2018.

\bibitem{krish}
Savitri Krishnamurthy, Stuart~J. Schnitt, Anne Vincent-Salomon, Rita
  Canas-Marques, Eugenia Colon, Kanchan Kantekure, Marina Maklakovski, Wilfrid
  Finck, Jeanne Thomassin, Yuval Globerson, Lilach Bien, Giuseppe Mallel, Maya
  Grinwald, Chaim Linhart, Judith Sandbank, and Manuela Vecsler.
\newblock Fully automated artificial intelligence solution for human epidermal
  growth factor receptor 2 immunohistochemistry scoring in breast cancer: A
  multireader study.
\newblock {\em JCO Precision Oncology}, page e2400353, 2024.
\newblock PMID:.

\bibitem{saha2018her2net}
Monjoy Saha and Chandan Chakraborty.
\newblock Her2net: A deep framework for semantic segmentation and
  classification of cell membranes and nuclei in breast cancer evaluation.
\newblock {\em IEEE Transactions on Image Processing}, 27(5):2189--2200, 2018.

\bibitem{cordeiro2018automatic}
Caroline~Q Cordeiro, Sergio~O Ioshii, Jeovane~H Alves, and Lucas~F Oliveira.
\newblock An automatic patch-based approach for her-2 scoring in
  immunohistochemical breast cancer images using color features.
\newblock {\em arXiv preprint arXiv:1805.05392}, 2018.

\bibitem{kabakcci2021automated}
Kaan~Aykut Kabak{\c{c}}{\i}, Asl{\i} {\c{C}}ak{\i}r, {\.I}lknur T{\"u}rkmen,
  Beh{\c{c}}et~U{\u{g}}ur T{\"o}reyin, and Abdulkerim {\c{C}}apar.
\newblock Automated scoring of cerbb2/her2 receptors using histogram based
  analysis of immunohistochemistry breast cancer tissue images.
\newblock {\em Biomedical Signal Processing and Control}, 69:102924, 2021.

\bibitem{WU2023100054}
Si~Wu, Meng Yue, Jun Zhang, Xiaoxian Li, Zaibo Li, Huina Zhang, Xinran Wang,
  Xiao Han, Lijing Cai, Jiuyan Shang, Zhanli Jia, Xiaoxiao Wang, Jinze Li, and
  Yueping Liu.
\newblock The role of artificial intelligence in accurate interpretation of
  her2 immunohistochemical scores 0 and 1+ in breast cancer.
\newblock {\em Modern Pathology}, 36(3):100054, 2023.

\bibitem{tewary2022autoihcnet}
Suman Tewary and Sudipta Mukhopadhyay.
\newblock Autoihcnet: Cnn architecture and decision fusion for automated her2
  scoring.
\newblock {\em Applied Soft Computing}, 119:108572, 2022.

\bibitem{kabir2024utility}
Saidul Kabir, Semir Vranic, Rafif~Mahmood Al~Saady, Muhammad~Salman Khan, Rusab
  Sarmun, Abdulrahman Alqahtani, Tariq~O Abbas, and Muhammad~EH Chowdhury.
\newblock The utility of a deep learning-based approach in her-2/neu assessment
  in breast cancer.
\newblock {\em Expert Systems with Applications}, 238:122051, 2024.

\bibitem{chauhan2024multiple}
Ekansh Chauhan, Amit Sharma, Megha~S. Uppin, Manasa Kondamadugu, C.~V. Jawahar,
  and P.~K. Vinod.
\newblock Ipd-brain: An indian histopathology dataset for glioma subtype
  classification.
\newblock {\em Scientific Data}, 11(1):1403, Dec 2024.

\bibitem{jiao2024prediction}
Panpan Jiao, Qingyuan Zheng, Rui Yang, Xinmiao Ni, Jiejun Wu, Zhiyuan Chen, and
  Xiuheng Liu.
\newblock Prediction of her2 status based on deep learning in h\&e-stained
  histopathology images of bladder cancer.
\newblock {\em Biomedicines}, 12(7):1583, 2024.

\bibitem{chen2024towards}
Richard~J Chen, Tong Ding, Ming~Y Lu, Drew~FK Williamson, Guillaume Jaume,
  Andrew~H Song, Bowen Chen, Andrew Zhang, Daniel Shao, Muhammad Shaban, et~al.
\newblock Towards a general-purpose foundation model for computational
  pathology.
\newblock {\em Nature Medicine}, 30(3):850--862, 2024.

\bibitem{lu2021data}
Ming~Y Lu, Drew~FK Williamson, Tiffany~Y Chen, Richard~J Chen, Matteo Barbieri,
  and Faisal Mahmood.
\newblock Data-efficient and weakly supervised computational pathology on
  whole-slide images.
\newblock {\em Nature biomedical engineering}, 5(6):555--570, 2021.

\bibitem{Zhang_2022_CVPR}
Hongrun Zhang, Yanda Meng, Yitian Zhao, Yihong Qiao, Xiaoyun Yang, Sarah~E.
  Coupland, and Yalin Zheng.
\newblock Dtfd-mil: Double-tier feature distillation multiple instance learning
  for histopathology whole slide image classification.
\newblock In {\em Proceedings of the IEEE/CVF Conference on Computer Vision and
  Pattern Recognition (CVPR)}, pages 18802--18812, June 2022.

\bibitem{liu2022convnet}
Zhuang Liu, Hanzi Mao, Chao-Yuan Wu, Christoph Feichtenhofer, Trevor Darrell,
  and Saining Xie.
\newblock A convnet for the 2020s.
\newblock In {\em Proceedings of the IEEE/CVF conference on computer vision and
  pattern recognition}, pages 11976--11986, 2022.

\end{thebibliography}

\section*{\centering{Supplementary}}

\setcounter{figure}{0} 

\renewcommand{\thefigure}{S\arabic{figure}} 

\setcounter{table}{0} 

\renewcommand{\thetable}{S\arabic{table}}

\begin{figure}[h!]
    \centering
    \includegraphics[width=\textwidth]{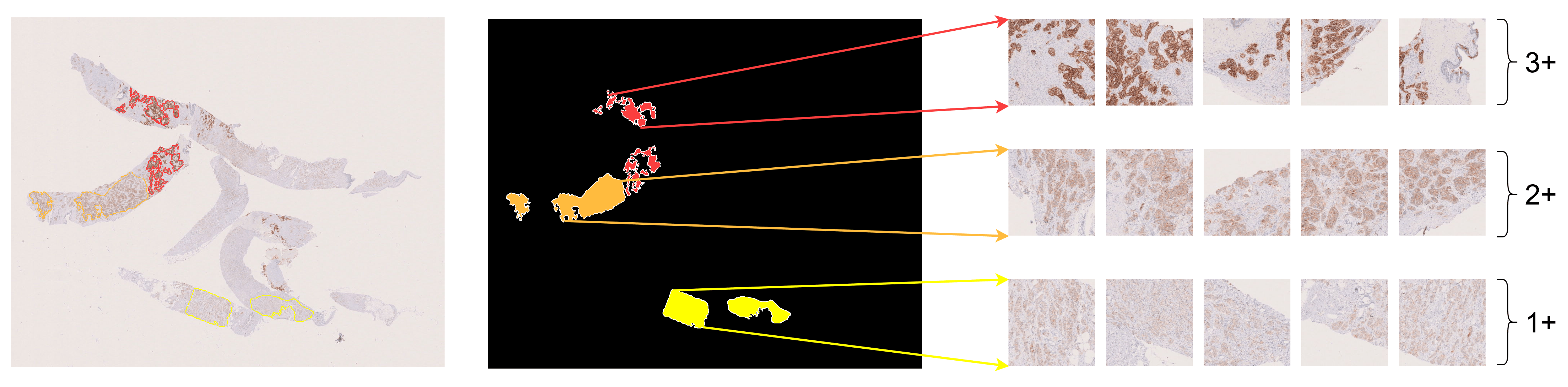}
    \caption{Sample conversion from a slide to patches with annotations for various HER2 scores. The left image shows the annotated slide, the middle image displays the extracted regions for each HER2 score, and the right image presents example patches categorized as 1+, 2+, and 3+ based on their HER2 scores in given annotation.}
    \label{fig:patch_dataset}
\end{figure}

\begin{table}[ht]
    \centering
    \begin{tabular}{|c|c|c|c|c|}
        \hline
        \textbf{Task} & \textbf{Method} & \textbf{AUC (\%)} & \textbf{F1 (\%)} & \textbf{Accuracy (\%)} \\
        \hline
        \multirow{3}{*}{Binary} & \textbf{CLAM-sb} & 96.98 $\pm$ 2.91 & 87.25 $\pm$ 6.70 & 89.6 $\pm$ 5.56 \\
        \cline{2-5}
         & \textbf{CLAM-mb} & \textbf{96.15 $\pm$ 3.21} & \textbf{88.45 $\pm$ 6.40} & \textbf{90.6 $\pm$ 5.58} \\
        \cline{2-5}
         & DTFD & 96.41 $\pm$ 0.15  & 87.15 $\pm$ 2.18  & 91.6 $\pm$ 1.45 \\
        \hline
        \hline
        \multirow{3}{*}{3-Way} & \textbf{CLAM-sb} & \textbf{92.05 $\pm$ 3.97} & \textbf{77.67 $\pm$ 8.65} & \textbf{78.6 $\pm$ 7.60} \\
        \cline{2-5}
         & CLAM-mb & 91.81 $\pm$ 3.04 & 76.15 $\pm$ 6.23 & 76.6 $\pm$ 6.73 \\
        \cline{2-5}
         & DTFD & 91.64 $\pm$ 0.58 & 76.48 $\pm$ 2.30 & 76.7 $\pm$ 2.31  \\
        \hline
        \hline
        \multirow{3}{*}{4-Way} & CLAM-sb & 92.66 $\pm$ 2.32 & 74.37 $\pm$ 5.41 & 74.4 $\pm$ 5.14 \\
        \cline{2-5}
         & CLAM-mb & 91.8 $\pm$ 2.22 & 73.55 $\pm$ 5.72 & 73.6 $\pm$ 5.15 \\
        \cline{2-5}
         & \textbf{DTFD} & \textbf{92.79 $\pm$ 0.95 } & \textbf{75.72 $\pm$ 1.61 } & \textbf{75.5 $\pm$ 1.58 } \\
        \hline
    \end{tabular}
    \caption{Approach-1 (MIL) Results: 10-Fold Performance of Various Attention-Based Classifiers on Test Set for Slide-Level Classification Tasks.}
    \label{tab:MIL_results}
\end{table}

\begin{table}[ht]
    \centering
    \begin{tabular}{|c|c|c|c|c|}
        \hline
        \textbf{Task} & \textbf{Method} & \textbf{AUC (\%)} & \textbf{F1 (\%)} & \textbf{Accuracy (\%)} \\
        \hline
        \multirow{3}{*}{Binary} & \textbf{Random Forest} & 96.33 $\pm$ 3.15 & \textbf{89.98 $\pm$ 5.37} & \textbf{92.20 $\pm$ 4.18} \\
        \cline{2-5}
         & SVM & 96.49 $\pm$ 2.90 & 89.92 $\pm$ 5.44 & \textbf{92.20 $\pm$ 4.18} \\
        \cline{2-5}
         & MLP & \textbf{97.36 $\pm$ 2.62} & 89.92 $\pm$ 5.44 & \textbf{92.20 $\pm$ 4.18} \\
        \hline
        \hline
        \multirow{3}{*}{3-Way} & \textbf{Random Forest} & 91.88 $\pm$ 3.44 & \textbf{78.11 $\pm$ 6.86} & \textbf{79.80 $\pm$ 6.55} \\
        \cline{2-5}
         & SVM & 91.83 $\pm$ 3.56 & 76.53 $\pm$ 6.36 & 78.32 $\pm$ 5.98 \\
        \cline{2-5}
         & MLP & \textbf{92.38 $\pm$ 3.27} & 76.84 $\pm$ 4.76 & 78.17 $\pm$ 5.05 \\
        \hline
        \hline
        \multirow{3}{*}{4-Way} & Majority Voting & 51.81  & 17.24 & 26.32 \\
        \cline{2-5}
         & \textbf{Random Forest} & \textbf{92.59 $\pm$ 2.73}  & \textbf{73.86 $\pm$ 6.75} & \textbf{74.48 $\pm$ 6.72} \\
        \cline{2-5}
         & SVM & 92.54 $\pm$ 2.84 & 73.65 $\pm$ 5.88 & 74.33 $\pm$ 5.81 \\
        \cline{2-5}
         & MLP & 92.43 $\pm$ 3.14 & 73.28 $\pm$ 5.48 & 73.65 $\pm$ 5.93 \\
        \hline
    \end{tabular}
    \caption{Approach-3 (Patch-Based) Results at \textbf{Moderate Resolution}: 10-Fold Performance of Various Machine Learning Classifiers on Test Set for All Slide-Level Classification Tasks Using Count of HER2 Score Patches.}
    \label{tab:high_res_classifier}
\end{table}

\begin{table}[ht]
    \centering
    \begin{tabular}{|c|c|c|c|c|}
        \hline
        \textbf{Task} & \textbf{Method} & \textbf{AUC (\%)} & \textbf{F1 (\%)} & \textbf{Accuracy (\%)} \\
        \hline
        \multirow{3}{*}{Binary} & \textbf{Random Forest} & 95.88 $\pm$ 3.77 & \textbf{89.50 $\pm$ 7.40} & \textbf{92.13 $\pm$ 5.27} \\
        \cline{2-5}
         & SVM & 95.13 $\pm$ 5.24 & 88.52 $\pm$ 9.36 & 91.78 $\pm$ 5.93 \\
        \cline{2-5}
         & MLP & \textbf{96.67 $\pm$ 3.33} & 88.52 $\pm$ 9.36 & 91.78 $\pm$ 5.93 \\
        \hline
        \hline
        \multirow{3}{*}{3-Way} & Random Forest & 89.79 $\pm$ 4.41 & 74.00 $\pm$ 8.46 & 75.94 $\pm$ 8.33 \\
        \cline{2-5}
         & \textbf{SVM} & 90.66 $\pm$ 3.57 & \textbf{75.69 $\pm$ 7.31} & \textbf{77.43 $\pm$ 7.29} \\
        \cline{2-5}
         & MLP & \textbf{91.15 $\pm$ 3.36} & 74.53 $\pm$ 7.81 & 76.55 $\pm$ 7.38 \\
        \hline
        \hline
        \multirow{3}{*}{4-Way} & Majority Voting & 42.24 & 21.05 & 30.77 \\
        \cline{2-5}
         & Random Forest & 91.29 $\pm$ 3.01  & 72.12 $\pm$ 5.94 & 72.82 $\pm$ 5.57 \\
        \cline{2-5}
         & \textbf{SVM} & 91.30 $\pm$ 2.79 & \textbf{73.26 $\pm$ 5.83} & \textbf{73.84 $\pm$ 6.21} \\
        \cline{2-5}
         & MLP & \textbf{91.85 $\pm$ 2.92} & 73.15 $\pm$ 5.84 & 73.55 $\pm$ 6.08 \\
        \hline
    \end{tabular}
    \caption{Approach-3 (Patch-Based) Results at \textbf{High Resolution}: 10-Fold Performance of Various Machine Learning Classifiers on Test Set for All Slide-Level Classification Tasks Using Count of HER2 Score Patches.}
    \label{tab:very_high_res_classifier}
\end{table}

\end{document}